\documentclass{article} 
\pdfoutput=1
\usepackage{nips15submit_e,times}
\usepackage{url}

\usepackage[utf8]{inputenc}
\usepackage{amsmath}
\usepackage{amssymb}
\usepackage{booktabs}
\usepackage{flushend}
\usepackage{multirow}
\usepackage{tabularx}
\usepackage{cite}

\usepackage[pdftex]{graphicx}
\usepackage{epstopdf}
   \graphicspath{{../images/}}
   \DeclareGraphicsExtensions{.pdf,.jpeg,.png,.eps}

\title{Feature Learning from Spectrograms\\ for Assessment of Personality Traits}

\author{
Marc-André Carbonneau\thanks{Corresponding author{marcandre.carbonneau@gmail.com} }, Eric~Granger, Yazid~Attabi \&  Ghyslain~Gagnon\\
Laboratoire d'imagerie, de vision et d'intelligence artificielle\\
École de technologie supérieure\\
Montreal, Canada\\
}

\nipsfinalcopy 

\begin{document}

\maketitle

\begin{abstract}
Several methods have recently been proposed to analyze speech and automatically infer the personality of the speaker. These methods often rely on prosodic and other hand crafted speech processing features extracted with off-the-shelf toolboxes. To achieve high accuracy, numerous features are typically extracted using complex and highly parameterized algorithms. In this paper, a new method based on feature learning and spectrogram analysis is proposed to simplify the feature extraction process while maintaining a high level of accuracy.
The proposed method learns a dictionary of discriminant features from patches extracted in the spectrogram representations of training speech segments. Each speech segment is then encoded using the dictionary, and the resulting feature set is used to perform classification of personality traits. Experiments indicate that the proposed method achieves state-of-the-art results with a significant reduction in complexity when compared to the most recent reference methods. The number of features, and difficulties linked to the feature extraction process are greatly reduced as only one type of descriptors is used, for which the 6 parameters can be tuned automatically. In contrast, the simplest reference method uses 4 types of descriptors to which 6 functionals are applied, resulting in over 20 parameters to be tuned.
\end{abstract}

\section{Introduction}

People spontaneously infer the personality of others from a wide range of cues. These cues may be visual, like facial expressions or posture, and may also be aural, like intonations, choice of words or voice timbre. This assessment of personality traits naturally influences the way we interact with each other \cite{Uleman1996}. The method proposed in this paper aims at performing this assessment automatically.


Being able to accurately predict the personality of an interlocutor is an important step toward better human-machine interactions. For example, people attribute personality traits to machines and interact differently with them depending on this perceived personality. For instance, extroverted people will interact longer with robots they perceive as extroverted \cite{Tapus2008}. Detecting and understanding a person's personality would enable a machine to adapt its behavior to the user. It can also be used in e-learning applications by giving appreciative feedback on the personality projected by a user to improve its leadership or sale skills.	 

In literature, five personality traits (the \textit{Big-Five}) corresponding to psychological phenomenon are observable regardless of the situation and culture: openness, conscientiousness, extroversion, agreeableness and neuroticism \cite{Digman1996}. These traits influence the way people act and speak. For instance, in \cite{Guadagno2008} a correlation is established between openness and neuroticism and the probability of maintaining blog. The choice of words by a subject based on his/her personality traits has also been studied in informal texts\cite{Argamon2005}, conversations \cite{Mairesse2007} and on social medias \cite{Qiu2012}. 

\label{Section:PersonalityDescription}

In the 2012 edition of the Interspeech competition on paralinguistics, one of the challenges was personality traits assessment from speech. This has motivated the proposition of several methods for this task. The baseline systems for the competition were designed using SVM and random forest (RF) classifiers trained with 6125-dimensional feature vectors \cite{Interspeech2012}. They performed particularly well, and only two contestants were able to surpass their performance on the test set. It was observed that increasing the number of features tends to increase recognition performance \cite{Interspeech2012}, thus large feature sets were extracted in the hope of capturing more of the relevant discriminant information. Some of the features were redundant or non-informative which motivated some contestants to use feature selection on the set of 6125 features \cite{Chastagnol2012,Wu2012,Pohjalainen2012}. The winners of the competition \cite{Ivanov2012} added 21760 spectral features to the baseline feature set before performing selection.       

Since 2012, the Interspeech competition 6125-dimension feature set of the baseline system has grown even larger. In 2015, it had increased to 6373-dimension \cite{Steidl2015}. Many of these features are statistics on the usual prosody features such as pitch, formants and energy, as well as more complex features, such as log harmonics to noise ratio, harmonicity and psycho-acoustic spectral sharpness. All of these application specific feature extraction techniques require a fair knowledge and experience in speech processing to tune their parameters, select thresholds, pre-process data, etc. Moreover, results may vary significantly from one implementation to another which limits the reproductibility of the experiments. 

For instance, the RASTA \cite{Hermansky1994} algorithm was used to extract several features in the challenge. It involves the selection of filter coefficients, non-linear compression and expansion functions and their respective parameters \cite{Hermansky1994}. Also, several features related to local extrema were used, such as first and second order statistics on inter-maxima distance. The detection of these extrema necessitates a peak selection algorithm which must be tuned to achieve a high level of performance \cite{Schuller2015}. 

Many practitioners use software tools to extract prosody features, which accelerates the design of recognition solutions. However, even if these tools contain complete implementations of feature extraction algorithms, expertise in speech processing is required to configure the several parameters and options of each modules. For instance, in Praat \cite{Praat} there are 5 different methods for pitch extraction, each with 3 to 9 parameters to be set. In openSMILE \cite{openSmile}, one must choose between the cPitchACF (4 parameters) object and the cPitchShs object (9 parameters) to extract pitch, which in turn must be configured. The user may also use a pitch smoother, where four more parameters must be set. There are even more parameters to consider when extracting formants.

Aside from the complexity and variability of these feature extraction procedures, the use of large feature sets reduces the generalization capability of pattern recognition algorithms \cite{Eyben2016}. Indeed, larger feature space are subject to problems associated with the curse of dimensionality \cite{Bishop2006}. The exponential growth of the search space increases the amount of data needed to obtain a statistically significant representation of the data. Generally, in affective computing contexts, data is limited because collection is costly, which calls for more compact data representations. Moreover, smaller feature sets are desirable because they allow for faster training and classification.

The difficulties described above have been discussed by several researchers in the affective speech recognition community. The CEICES initiative attempted to create a standardized set of feature for emotion recognition in speech \cite{Batliner2006}. The proposed set is a combination of 381 acoustic and lexical features selected from a pool of 4024 features that the authors have successfully used in their previous research. While the collection of features was standardized, the implementation of the feature extraction algorithms was not. Recently, another attempt has been made to reduce the size of the feature collection used for automatic voice analysis \cite{Eyben2016}. A minimal number of descriptors were selected based on theoretical and empirical evidence. While the minimal and extended sets are compact (62 and 88 features respectively) several different algorithms are used for the extraction of the descriptors. Each algorithm requires expertise and a careful parametrization\footnote{The feature set has been made publicly available through the openSMILE toolkit \cite{openSmile}.}. 

In this paper, a method inspired by the recent developments in feature learning and image classification is proposed to alleviate these design choices for automatic assessment of personality traits. The temporal speech signals are translated into spectrogram images. Small sub-images, called patches, are densely extracted from these spectrogram images, and used during training to learn a feature dictionary yielding a sparse representation. The dictionary is used to encode each of the local patches. Each spectrogram is thus represented as a collection of encoded patches, which are pooled to create a histogram representation of the entire spectrogram. These histograms are used to train a classifier. During testing, a new speech signal is represented by a histogram, using the same dictionary, before classification. 

The proposed method of representation, which is based on local patches, allows to capture para-linguistic information compactly. Because it encodes raw parts of the spectrogram images, the representation is richer than methods which characterize speech signals with statistics on the whole signal \cite{Mohammadi2012,Eyben2016,Interspeech2012}. For instance, these methods use the mean, the standard deviation, kurtosis, min and max of the pitch or spectrum and cepstrum bins, which discard the relevant cues for personality assessment that the local shape of the signal contains. Moreover, when compared to these methods, the proposed method has fewer parameters, which can be more easily tuned using standard automatic hyper-parameter optimization techniques (e.g. cross-validation). Finally, the method inherits the robustness to deformation and noise of local image recognition methods applied to spectrogram analysis \cite{Schutte2009,Sharan2015}.

In essence, the proposed method leverages the power of representation inherent to sparse modeling, which learn features from the data. This approach generally leads to a high level of accuracy \cite{Grosse2007}. The dimensionality of feature vectors needed for this level of performance is reduced by an order of magnitude when compared to the number of features used in the Interspeech challenges. Moreover, only one method is used for feature extraction which limits the number of parameters needing careful tuning. Finally, the proposed technique does not necessitate a feature selection stage which is usually time consuming during training. 

The proposed method is compared to 6 reference methods on the SSPNet Speaker Personality Corpus used in the Interspeech 2012 competition. As stated in the \textit{postmortem} report of the challenge published in 2015 \cite{Schuller2015}, research in automated recognition of speaker traits is still active, and still requires much exploration to isolate suitable features and models for this task. In this regard, the novel technique proposed in this paper aims to provide a simpler alternative for extraction of a compact set of features that achieve state-of-the-art results.

The rest of the paper is organized as follows: The next section provides background information on feature learning in the context of speech analysis. Section \ref{Section:Proposed} describes the proposed method. Section \ref{Section:Exp} presents the experimental data, protocol and reference methods. The results are analyzed in Section \ref{Section:Res}.

\section{Feature Learning for Speech Analysis}
Feature learning algorithms extract relevant features themselves, instead of relying on man-engineered representations, which are time consuming to obtain and often sub-optimal. Feature learning has been used in several speech analysis applications. Some methods use deep neural networks, which intrinsically learn features, to perform automatic speech recognition (ASR) \cite{Morgan2012,Mohamed2012}. These systems are not suitable for personality trait recognition because they analyze local time series (e.g. a phoneme), and fail to capture the global information in a speech segment. Deep learning has also been used for automatic emotion recognition. In \cite{Kim2013}, a neural network learns a feature representation, not from the raw signal, but from a set of prosodic, spectral and video features. In \cite{Deng2013}, utterances where represented using a sparse auto-encoder to perform transfer learning in an emotion recognition task. In \cite{Heckmann2011}, base features were learned using ICA on spectrograms. After a feature selection process, the selected features were combined in a higher hierarchical level, using non-negative sparse coding. These feature combinations were used with an HMM to perform ASR.

Feature learning can be performed on several types of signal representation. When a speech signal is represented as a spectrogram, (i.e. concatenation in time of windowed Discrete Fourier Transform (DFT)), it can be analyzed through image processing. It has been demonstrated by neuroscientists that the same parts of the brain can be used to process both visual and audio signals \cite{vonMelchner2000}. This has motivated several researchers to investigate the application of image recognition techniques to spectrograms to analyze and recognize sound and speech signals. For example, histograms of oriented gradients (HOG) were used to perform word recognition \cite{Muroi2009}. In \cite{Dennis2014}, spectrograms amplitudes are quantized and mapped into a color coded image. Color distributions are then characterized and analyzed. This method is inspired by content-based image retrieval methods \cite{Shih2002}. In \cite{Sharan2015}, spectrograms and cochleograms are divided in frequency sub-bands and analyzed as visual textures using gray-tone spatial dependence matrix features \cite{Haralick1973} alongside cepstral features. Audio spectrograms were employed with a convolutional deep Bayesian network, typically used for image recognition, to perform speaker identification and gender classification \cite{Lee2009}. The representation achieved a higher recognition performance when compared to MFCC and raw spectrograms. The Gabor function (sinusoidal tapered by a decaying exponential), were found to be good models of receptive fields in the human visual cortex \cite{Marcelja80}. This has motivated several authors to apply log-Gabor filter banks to spectrograms\cite{Gu2015,Buisman2012} to analyze paralinguistics.


A popular paradigm for image analysis is to extract features locally (instead of globally) from salient regions of an image, called patches. The set of patches, is used to represent an entire image. This type of approach, often called bag-of-words, have been successfully applied in numerous contexts for recognition in image \cite{Philbin2007,Csurka2004} and video \cite{Laptev2008,Carbonneau2015}. Using local features in image recognition may lead to an increased robustness to intra-class variation, deformation, view-point, illumination and occlusion \cite{Zhang2006}. When working with spectrograms, it translates to an increased robustness to noise \cite{Schutte2009,Dennis2014}. In \cite{Matsui2011} the SIFT descriptor was used to detect and encode key-points in spectrogram images of musical pieces to perform genre classification. Schutte proposed a deformable part-based model of local spatio-temporal features in speech recognition \cite{Schutte2009}. The method allowed to improve recognition performance over the HMM baseline system especially in the presence of noise. 

Local-based methods in image recognition often exploit a set of predefined basis for decomposition such as wavelets, wedgelets and bandlets \cite{Mallat2008}. However, it has been shown that learning the basis directly on the data leads to a higher level of accuracy in several applications such as signal reconstruction \cite{Elad2006} and image classification \cite{Raina2007} and reconstruction \cite{Aharon2006}. Based on these results, several recently proposed spectrogram analysis methods learn representation on training data in order to benefit from the improved performance. For instance, in \cite{Lyon2010} the spectrograms are segmented at different scales, and each segment is encoded as the most resembling word in a dictionary learned using the \textit{k}-means algorithm. In \cite{Yu2009} the spectrograms of musical instruments are interpreted as visual textures. Sounds are represented by a vector encoding the resemblance between the spectrogram and a randomly constituted dictionary. 

In the aforementioned dictionary-based methods, local descriptors are associated with the most representative code-word in the dictionary. Some algorithms use sparse coding to perform this association and learn a representation \cite{Elad2006,Peyre2009}. Sparse coding is a type of feature learning which expresses a signal using a small number of basis from a learned set, usually called dictionary. Experiments have shown that encoding audio and visual signals using a sparse decomposition can lead to a high level of accuracy for various tasks such as speaker, gender and phoneme recognition \cite{Lee2009}. Also, it was shown that a learned sparse representation of audio signals is akin to the early mammalian auditory system \cite{Smith2006}. This is why several recent methods use sparse coding to learn the dictionary and encode signals. 

\begin{figure*}
\centering
\includegraphics[width=.9\textwidth]{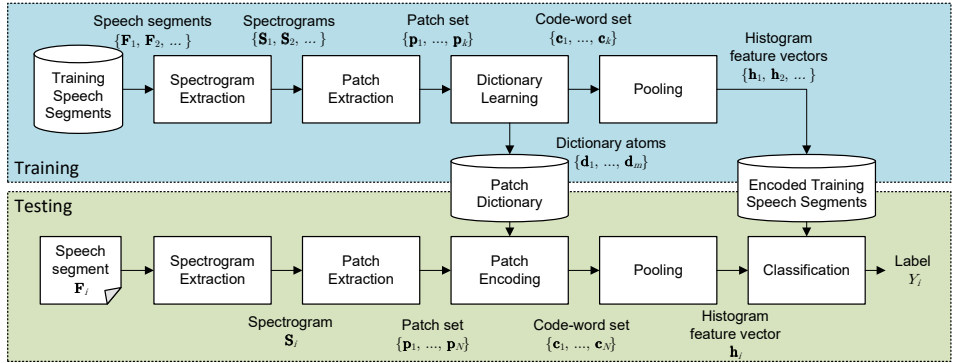} 
\caption{Block diagram of the proposed system for the prediction of a personality trait. The upper part illustrates the operations performed during training. The lower part illustrates sequence of operations performed to process an input speech sequence in test.}
\label{Fig:top}
\end{figure*}

\begin{figure*}
\centering
\includegraphics[width=\textwidth]{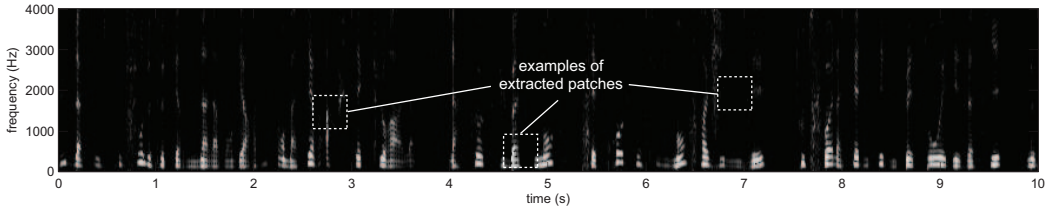} 
\caption{Example of spectrogram extracted from a speech file in the SSPNet database. White indicates high values while black indicates low values.}
\label{Fig:exampleSpect}
\end{figure*}

In the context of personality assessment from speech, paralinguistic cues have to be analyzed globally. Methods used in other speech analysis applications, such as ASR, fail to capture this global information. Existing methods for personality recognition capture global information using a statistical operators on low-level features. Unfortunately, this results in a high dimensional representation, which is prone to the curse of dimensionality, and require fair signal processing expertise to extract the low-level features. The proposed method represents a complete speech segment as an image then uses image recognition techniques, and thus, can perform global analysis. Moreover, it uses a feature learning approach, which reduces the burden associated with feature engineering and yields a compact representation, and leads to increased recognition performances.

\section{Proposed Feature Learning Method}
\label{Section:Proposed}
This section presents a new method for predicting personality traits in speech based on spectrogram analysis and feature learning. The main stages of the proposed method are depicted in Figure \ref{Fig:top}.  Specifics details regarding our proposed solution for feature extraction, classification and dictionary learning are described in the next sections. The upper part is the pipeline for training. At first, for each speech segment $\textbf{F}$ in the data set, a spectrogram $\textbf{S}$ is extracted by applying a Fourier transform on a sliding window, yielding a 2-dimensional matrix. Small sub-matrices, called patches $\{\textbf{p}_1, ..., \textbf{p}_k \}$ are then uniformly extracted from all the spectrogram matrices in the training set. A dictionary $\textbf{D} = \{\textbf{d}_1, ..., \textbf{d}_m\}$ is learned from these patches, and at the same time, the patches are encoded as sparse vectors called code-words $\{\textbf{c}_1, ..., \textbf{c}_k \}$. A single $m$-dimensional feature vector representation $\textbf{h}$ is obtained for each training speech sample by pooling together all code-words extracted from it. A two-class support vector machine (SVM) classifier is trained using these feature vectors for each personality trait. 

The lower part is the pipeline used during testing, to predict a personality trait. Like in training, patches are extracted from the spectrograms. Each patch is encoded using the previously learned dictionary. The resulting code-words are then pooled to create a feature vector that is fed to a 2-class classifier to obtain a label $Y$ representing to which end of the spectrum of a specific personality trait the speech segment corresponds.

\subsection{Feature Extraction}
Given a speech segment $x(n)$, the spectrogram $\mathbf{S}$  is the concatenation in time of its windowed DFT: 
\begin{equation}
\mathbf{S} = \{\mathbf{X}_0, ..., \mathbf{X}_t, ..., \mathbf{X}_T\},
\end{equation}
where $\mathbf{X}_t$ is a column vector containing the absolute amplitude of the DFT frequency bins and $T$ is the number of DFTs extracted from the signal. The absolute amplitude is favored over the log-amplitude as it has shown to yield better results for spectrogram image classification in\cite{Dennis2014} and in our own experiments. The spectrograms are normalized: each frequency bin is divided by the maximum amplitude value contained in a time frame. This process results in a 2-D matrix $\mathbf{S}$ which can be analyzed as a grey-scale image. An example of spectrogram extracted on the SSPNet Speaker Personality Corpus is illustrated in Figure \ref{Fig:exampleSpect}. 

From the matrix $\mathbf{S}$, small patches, or sub-images, of $p \times p$ pixels are extracted at regular intervals. A vector representation $\mathbf{p}_i \in \mathbb{R}^{1 \times d}$ of each patch ($d=p \times p$) is obtained by concatenating the value of all pixels. The vector $\mathbf{p}_i$ is encoded into $\mathbf{c}_i$ using a previously learned dictionary $\mathbf{D}$ containing $m$ atoms (more details in Section \ref{Section:DL}). These atoms are vector basis that are used to reconstruct the patches. The code-vector $\mathbf{c}_i$ corresponding to the patch $\mathbf{p}_i$ is obtained by solving
\begin{equation}
l(\mathbf{c}_i) \triangleq \underset{\mathbf{c}_i \in \mathbb{R}^{d}}{\text{min}}\; \frac{1}{2}\left \| \mathbf{p_i}-\mathbf{Dc}_i \right \|^2_2 + \lambda\left \| \mathbf{c}_i \right \|_1
\label{Equation:coding}
\end{equation}
using the LARS-Lasso algorithm \cite{Efron2004}. The loss function has two terms, each encoding an optimization objective, and $\lambda$ is a parameter used to adjust the relative importance of the two terms. The first term is the quadratic reconstruction error, while in the second term, the $\ell_1$ norm of the code vector is used to enforce sparseness.
Once a code $\mathbf{c}_i$ is obtained for each patch $\mathbf{p}_i$, the absolute value of all the codes are summed to obtain a histogram $\mathbf{h}$ describing the entire spectrogram $\mathbf{S}$:
\begin{equation}
\mathbf{h} = \sum_{i} \left | \mathbf{c}_i \right |
\end{equation} 
These histograms represent the distribution of patches over speech segments. It is thus possible to directly compare segments of different length.  

\subsection{Classification}
The speech segments are represented by histograms and thus, appropriate distance measure should be employed. Several distance measures have been proposed to compare histograms. In this paper's implementation, the $\chi^2$ distance is used because it showed competitive performances for visual bag-of-words histograms \cite{Zhang2006}. The $\chi^2$ distance is given by :
\begin{equation}
d_{\chi^2}(\mathbf{g},\mathbf{h}) = \sum_{i=1}^{m}\frac{(g_{i}-h_{i})^2}{g_{i}+h_{i}},
\label{Eq:dist}
\end{equation}
where  $g_i$ and $h_i$ are the $i^{\text{th}}$ bins of histograms \textbf{h} and \textbf{y}, and \(m\) corresponds to the number of words in the dictionary.

In this paper $d_{\chi^2}$ is used in an SVM framework with an exponential kernel\cite{Chapelle1999}:
\begin{equation}
\label{chi}
k(\textbf{g},\mathbf{h}) = e^{-\gamma d_{\chi^2}(\mathbf{g},\mathbf{h})},
\end{equation}
where the parameter \(\gamma\) controls the kernel size. 

While the implementation of this paper employs the $\chi^2$ distance and an SVM classifier, the proposed methods is not bound to these choices, and other distance functions and classifiers can be used.

\subsection{Dictionary Learning}
\label{Section:DL}
The objective of the dictionary learning phase is to generate a representative dictionary $\mathbf{D}=\left [ \mathbf{d}_1, ..., \mathbf{d}_m  \right ]  \in \mathbb{R}^{d \times m}$ given the matrix $\mathbf{P} = \left [ \mathbf{p}_1, ..., \mathbf{p}_k  \right ] \in \mathbb{R}^{d \times k}$ containing patch vectors extracted from the training set. Generally, for image classification tasks, best results are obtained with over-complete ($m > d$) dictionaries \cite{Tosic2011}. 

A dictionary of atoms $\mathbf{D}$ and sparse code-words $\mathbf{C}$ can be obtained by minimizing the following loss function: 
\begin{equation}
l(\mathbf{C},\mathbf{D}) \triangleq \underset{\mathbf{C}\in \mathbb{R}^{m \times k},\mathbf{D \in \mathcal{C}}}{\text{min}}\; \frac{1}{2}\left \| \mathbf{P}-\mathbf{DC} \right \|^2_2 + \lambda\left \| \mathbf{C} \right \|_1
\end{equation}
In this equation, $\lambda$ is the same as in (\ref{Equation:coding}) and is used to adjust the weight of the sparseness term in the loss equation. The convex set:
\begin{equation}
\begin{aligned}
\mathcal{C} \triangleq \{ \mathbf{D}\in\mathbb{R}^{d \times m} \; &  \textbf{s.t.} \; \forall i=1,...,m, \mathbf{d}^T_i\mathbf{d}_i\ \leq 1 \\  &\textbf{and} \; \forall i=1,...,m, \mathbf{d}_{i} \in \mathbb{R}_{\geq0} \}
\end{aligned}
\end{equation}
enforces two constraints. The first is used to restrict the magnitude of the dictionary atoms. The second is used to make sure each element of each atom in the dictionary is positive. Since the spectrogram is purely positive, better results are obtained by enforcing this constraint.
The joint optimization of $\mathbf{C}$ and $\mathbf{D}$ is not convex. However if one term is fixed the problem becomes convex. Thus, a common  strategy is two alternate between updating $\mathbf{C}$ while $\mathbf{D}$ is fixed and updating $\mathbf{D}$ while $\mathbf{C}$ is fixed until a stopping criterion is met \cite{Lee2006}. 

\begin{figure}
\centering
\includegraphics[width=0.42\textwidth]{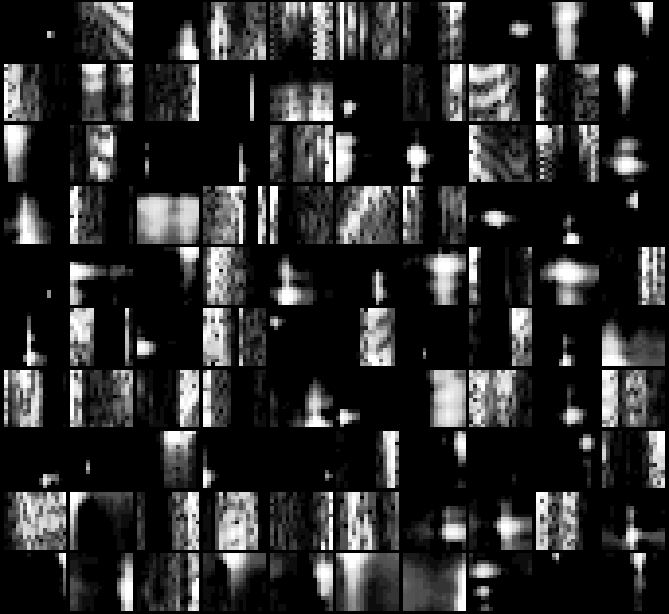} 
\caption{Example of patches from a 100 word dictionary created with sparse coding.}
\label{Fig:exampleDict}
\end{figure}

Figure \ref{Fig:exampleDict} shows an example of dictionary atoms learned using the above described procedure. The audio files from the SSPNet Speaker Personality Corpus were used to learn the atoms. The same dictionary can be used for all traits.

\section{Experimental Methodology}
\label{Section:Exp}

The SSPNet Speaker Personality corpus \cite{Mohammadi2012} is the largest and most recent data set for personality trait assessment from speech. It consists of 640 audio clips randomly extracted from french news bulletins in Switzerland. All clips have been sampled at 8 kHz and most of the clips are 10 seconds long, but some are shorter. Each clip contains only one of the 322 different speakers. Eleven judges performed annotation on each clip by completing the BFI-10 personality assessment questionnaire \cite{Rammstedt2007}. From the questionnaire a score is computed for each of the \textit{Big-Five} personality traits. Precautions were taken to avoid sequence and tiredness effects in the annotation process. The judges did not understand french and therefore were not influenced by linguistic cues. 
In \cite{Mohammadi2012} the assessment of the judges were considered as positive if the score was greater than 0 and negative otherwise. The labeling scheme was refined for the competition \cite{Interspeech2012}. In this case, an assessment was considered positive if the score given by a judge was higher than the average score given by this particular judge for the trait. In both cases, the final label for an instance was obtained by a majority vote from all of the 11 judges. Preliminary experiments showed a 1$\sim$2\% difference in accuracy performance between the two labeling schemes. The results reported in this paper were obtained using the competition's labeling scheme.

\setlength{\extrarowheight}{0pt}
\begin{table*}
\caption{Performance on the SSPNet Speaker Personality corpus and parameter complexity of the methods.}
\center
\scalebox{0.70}{
\begin{tabular}{lcccccccccc}
\toprule
& \multicolumn{6}{c}{\textbf{Unweighted Average Recall (\%)}} &\multicolumn{4}{c}{\textbf{Number of}} \\
\cmidrule(r){2-7}
\cmidrule(r){8-11}
\textbf{Algorithm} 
&O & C & E & A & N & Avr. & Features & Descriptors & Functionals & Parameters \\

\midrule
Mohammadi \& Vinciarelli (LR) \cite{Mohammadi2012}
& 56.1	& 69.6	& 72.4	& 55.7	& 67.4	& 64.2
& 24 & 4 & 6 & $>$20
\\
Mohammadi \& Vinciarelli (SVM) \cite{Mohammadi2012}
& 57.7	& 68.0	& 74.3	& 57.4	& 65.5	& 64.6
& 24 & 4 & 6 & $>$20
\\

SSPNet Challenge Baseline (SVM) \cite{Interspeech2012} 
& \textbf{58.7} 	& 69.2 	& 74.5	& 62.2	& 69.0 	& 66.7 
& 6125 & 21 & 39 & $>$200
\\
SSPNet Challenge Baseline (RF) \cite{Interspeech2012} 
& 52.9	& 69.0	& \textbf{77.5}	& 60.1	& 68.2	& 65.5
& 6125 & 21 & 39 & $>$200
\\

GeMAPS (SVM) \cite{Eyben2016}
& 56.3	& 72.2	& 74.9	& 61.9	& 68.9	& 66.8
& 62 & 13 & 10 & $>$100
\\
eGeMAPS (SVM) \cite{Eyben2016}
& 53.7 & \textbf{72.5}	& 75.1	& 62.0	& 66.6 & 66.0
& 88 & 16 & 12 & $>$100
\\
\midrule
Proposed Method 
& 56.3	& 68.3	& 75.2	& \textbf{64.9}	& \textbf{70.8} & \textbf{67.1} 
& 200-800 &1 & 1& 6
\\
\bottomrule
\end{tabular}}
\label{Table:ResultsPersonality}
\end{table*}

The metric used to compare accuracy is the unweighted average recall (UAR), which is the same as in the competition. The UAR is the mean of each class accuracy, and thus is unaffected by class imbalance. To assess performances, a 3-fold cross-validation procedure was used to limit the effect of sampling-induced variance in the results. Precautions were taken to make sure that all samples belonging to the same speaker are grouped in the same fold. Sampling-induced variance effects were observed in the Interspeech 2012 challenge. The results obtained for the conscientiousness trait with the development partition are significantly lower than the results obtained with the test partition. For instance, the baseline method using SVM obtained a UAR of 74.5\% in training, but increased to 80.1\% in testing \cite{Interspeech2012}. The same phenomenon was observed with the random forest classifier (74.9\% to 79.1\%). This suggests that the test data may have been easier to classify than the average data. This hypothesis is supported by the fact that the results obtained using a cross-validation procedure in \cite{Mohammadi2012} were also closer to 70\% than 80\%.  
Nested cross-validation \cite{stone1974cross} was used to optimize the hyper parameters for all classifiers and the dictionary learning parameters (dictionary size and $\lambda$). In nested cross-validation, an outer cross-validation loop (3 folds) is used to obtain the final test results, and an inner loop (5 folds) is used to find the best hyper parameter via grid search. Hyper-parameter optimization is thus performed for each of the 3 test folds separately.



For the proposed method, the spectrograms were extracted using a STFT with 128 sample wide Hamming window. There was a 75\% overlap between two successive speech segments. The extracted patches were 16$\times$16 \textit{pixels} thus yielding 256-dimensional feature vectors. A new patch was extracted each 8 time steps and each 4 frequency bins. All of these parameters were selected based on prior experiments. An importance weighting scheme was used to deal with class imbalance \cite{Rosenberg2012}. This was achieved by attributing different misclassification cost in the SVM hinge loss function to the target classes. The cost for the positive class was multiplied by a factor corresponding to the class imbalance ratio. 
The SPAMS toolbox \cite{Mairal2009} was used for dictionary learning and encoding and LIBSVM \cite{LIBSVM} was used for the SVM implementation. 
Three reference methods were selected to compare performance. The methods were chosen because they are well documented and can be reproduced without ambiguity. The first method was proposed by Mohammadi \& Vinciarelli in \cite{Mohammadi2012}. Prosody features were extracted using Praat \cite{Praat}, the same software used in the original paper. The low-level feature extracted were pitch, first two formants, energy of speech, and length of voiced and unvoiced segments. The features were extracted using 40 ms long windows at 10 ms time steps. The features were whitened based on means and standard deviations estimated on the training folds. Four statistical properties were then estimated from the 6 prosody measures yielding a 24-dimensional feature vector for each speech file. The statistical features were the minimum, maximum, mean and the entropy of the differences between consecutive feature values. As in \cite{Mohammadi2012}, an SVM and a logistic regression (LR) were used for classification. The logistic regression implementation of the MATLAB Statistic and Machine Learning Toolbox was used. For the SVM, the LIBSVM implementation was used with the linear and the RBF kernels. 

The second method is the baseline used in the Interspeech 2012 speaker trait challenge \cite{Interspeech2012}. The 6125 low-level features were extracted using the openSMILE software \cite{openSmile} with the preset named after the challenge. The features were whitened based on means and standard deviations estimated on the training folds. For the linear SVM, the LIBSVM implementation \cite{LIBSVM} was used which performs sequential minimal optimization, the optimization algorithm used in the challenge baseline. 
The use of Gaussian kernel was also explored but did not yield better results. For the random forest (RF) classifier, MATLAB implementation from the Statistic and Machine Learning Toolbox was used. 
This method was selected because it yield state-of-the-art performance. Only 2 of the methods proposed in the challenge outperformed the baseline with a UAR margin of 0.1\% for \cite{Montacie2012} and of 1\% for \cite{Ivanov2012}, which is not significant. 

The third and most recent benchmark method uses the features prescribed in the Geneva minimalistic acoustic parameter set (GeMAPS) \cite{Eyben2016}. The minimalistic set can be extended (eGeMAPS) by including MFCC coefficients, spectral flux and additional formant descriptors. The features were extracted using the preset supplied in openSMILE. Classification was achieved by a linear SVM using the LIBSVM implementation. The hyper-parameters were optimized in the same way as for the Interspeech method. This method was selected because it is intended to reduce the complexity of the feature extraction stage in paralinguistic problems, same as the proposed method.

\section{Results}
\subsection{Accuracy}
\label{Section:Res}

The performance of the proposed and baseline methods on the SSPNet Speaker Personality data set is reported in Table \ref{Table:ResultsPersonality}. The best average UAR was obtained using the proposed method. However, the results obtained when using the challenge features and GeMAPS with an SVM classifier are comparable. The method proposed by Mohammadi and Vinciarelli yields slightly lower accuracy than the other methods, although the difference in performance in most cases is small and may be insignificant. Particularities in the data set and the type of classifier, as well as its implementation, are most likely the reason for these variations in performance. For instance, using the same features and a different classifier, the SSPNet challenge baseline \cite{Interspeech2012} obtains a UAR of 58.7\% (SVM) and 52.9\% (RF) for the openness trait. The difference in UAR is not due to the power of the data representation, but on the behavior of the classifiers in this particular case. 

There are, however, differences between the four representations. For instance, the proposed method is not well adapted to represent pitch nor speech rate. Estimating the pitch is difficult because once the patches are extracted, their location is discarded. In contrast, all reference methods explicitly extract pitch and compute statistics on the measure. Speech rate is also difficult to represent by the proposed method since patches encode local information while speech rate is more of a global measure. All reference methods capture speech rate better because they extract statistics on the length and proportion of voiced and unvoiced segments. This slightly impedes the proposed method for the recognition of the openness trait, for which pitch and speech rate have been identified as markers \cite{Mairesse2007,Addington1968}. It could explain the 2.4\% and 1.4\% difference between the proposed and reference methods using SVM. However, these two markers are also indicative of neuroticism \cite{Mairesse2007}, and the proposed method performs better on this class because it can more efficiently capture voice timbre and prosody than the other methods. Instead of indirectly measuring it with formants, pitch and spectral features, the proposed method uses raw chunks of the sound spectrogram as representation, and thus might capture the information with more fidelity than binned descriptors and statistics over whole sound files.   
 
\subsection{Complexity}
While accuracy is generally similar for all methods, the main advantage of the proposed method is the significant reduction of complexity in implementation. Compared to the baseline of the Interspeech challenge, the feature set used in the proposed method is much smaller (at most 800 features instead of 6125). Smaller feature sets are desirable because they reduce algorithmic complexity, and are less subject to problems associated with the curse of dimensionality. Moreover, the amount of human expert intervention necessary is different in the four methods. In the proposed method, only 1 feature extraction algorithm was used instead of 4 for \cite{Mohammadi2012}, more than 10 for GeMAPS and over 20 in \cite{Interspeech2012}. In addition, in all reference methods, a set of functionals were applied to the extracted features. Some of these functionals were simple measures like mean, min/max and standard deviation, but other were more complex and parametrizable. For instance, functionals relying on peak distance need a peak detector that has to be fine-tuned. These feature extraction algorithms require parametrization which must be performed by a signal processing expert. 

The time complexity of the proposed method is slightly higher than for some of the other types of descriptors. This is due to the optimization step of the dictionary learning and encoding procedures. It takes under 3 seconds to analyze a 10 seconds speech segment using an unoptimized MATLAB implementation. Finally, one could argue that more memory is required with the proposed method as it needs to store the dictionary. However, a 800 word dictionary for 16$\times$16 \textit{pixel} patches require storing around 1.6 MB when using the double-precision floating-point format, which a manageable consumption in modern computers.







\section{Conclusion}
\label{Section:Concl}
This paper presents a new method for automated assessment of personality traits in speech. Speech segments are represented  using spectrograms and feature learning. The proposed representation is compact and is obtained using a single algorithm requiring minimal expert intervention, when compared to reference methods. Experiments conducted on SSPNet data set indicate that the proposed method yields the same level of accuracy as state-of-the-art methods in paralinguistics that employ more complex representations, while remaining simpler to use. 


As explained in Section \ref{Section:Res}, the method is not properly equipped to capture pitch and speech rate. Research should be conducted to include these signal characteristics in the representation. In addition, experiments on different paralinguistic problems should be conducted to validate the applicability of the proposed method in different contexts. Experiments should also be conducted where the sparse dictionary learning and classifier algorithms used in our implementation is replaced by other methods enforcing group sparsity and discrimination.

\bibliographystyle{IEEETran}
\bibliography{Personality}

\end{document}